\documentclass{article}

     \PassOptionsToPackage{numbers, compress}{natbib}

     \usepackage[preprint]{neurips_2025}

\usepackage[utf8]{inputenc} 
\usepackage[T1]{fontenc}    
\usepackage{hyperref}       
\usepackage{url}            
\usepackage{booktabs}       
\usepackage{amsfonts}       
\usepackage{nicefrac}       
\usepackage{microtype}      
\usepackage[dvipsnames]{xcolor}
\usepackage[ruled,linesnumbered,boxed,commentsnumbered]{algorithm2e}
\usepackage{adjustbox}
\usepackage{graphicx}
\usepackage{amssymb}
\usepackage{amsmath}
\usepackage{booktabs}
\usepackage{enumitem}
\usepackage{wrapfig}
\usepackage{tabularx}
\usepackage{pifont}
\usepackage{tcolorbox}
\usepackage{multirow}
\RestyleAlgo{ruled}
\SetKwComment{Comment}{// }{}
\DontPrintSemicolon
\newcommand{\Sref}[1]{\S\ref{#1}}
\mathchardef\mhyphen="2D
\newcolumntype{s}{>{\hsize=.1\hsize}X}
\newcolumntype{m}{>{\hsize=.3\hsize}X}
\newcommand{\cmark}{\ding{51}}
\newcommand{\xmark}{\ding{55}}
\definecolor{myblue}{HTML}{4A90E2}
\definecolor{mygreen}{HTML}{50C878}
\definecolor{myred}{HTML}{D9534F}
\definecolor{mygray}{rgb}{0.5,0.5,0.5}
\definecolor{myorange}{HTML}{FF7A33}

\newcommand{\ourmethod}[1]{\textsc{Data Swarms}}

\title{\ourmethod{}: Optimizable Generation of \\ Synthetic Evaluation Data}

\author{Shangbin Feng \ \ \ \ \ Yike Wang \ \ \ \ \ Weijia Shi \ \ \ \ \ Yulia Tsvetkov \\
University of Washington \\
\texttt{shangbin@cs.washington.edu}
}

\begin{document}

\maketitle

\vspace*{-20pt}
\begin{abstract}
  We propose \ourmethod{}, an algorithm to optimize the generation of synthetic evaluation data and advance quantitative desiderata of LLM evaluation. We first train a swarm of initial data generators using existing data, and define various \emph{evaluation objectives} to reflect the desired properties of evaluation (e.g., generate more \emph{difficult} problems for the evaluated models) and quantitatively evaluate data generators. We then employ particle swarm optimization to optimize the swarm of data generators, where they collaboratively search through the model parameter space to find new generators that advance these objectives. We further extend it to \textsc{Adversarial Swarms}, where the data generator swarm generates harder data while the test taker model swarm learns from such data, co-evolving dynamically for better data and models simultaneously. Extensive experiments demonstrate that \ourmethod{} outperforms eight data generation baselines across five evaluation objectives, while \textsc{Adversarial Swarms} produce more robust learning of synthetic data and stronger generalization. Further analysis reveals that \ourmethod{} successfully optimizes compositions of multiple evaluation objectives and generalizes to new off-the-shelf LLMs, unseen at optimization time.
\end{abstract}

\vspace*{-10pt}
\section{Introduction}

With new LLMs emerging daily---each trained with different training recipes, domain expertise, and alignment strategies---rigorous evaluation is more critical than ever to reveal the strengths and weaknesses of LLMs, and guide model selection for diverse applications. The most common evaluation approach relies on \emph{static} data, ranging from single-skill datasets \citep{zellers2019hellaswag, hendrycksmeasuring} to comprehensive multi-task benchmarks \citep{liangholistic, wangglue}. As state-of-the-art models become saturated on these static datasets \citep{open-llm-leaderboard-v2} with contamination concerns \citep{golchintime}, new evaluation methods focus on \emph{synthetic data generation} for incorporating temporal updates \citep{kasai2023realtime}, adaptively probing model weakness \citep{brown2025adaptively}, leveraging privileged information \citep{liautobencher}, and more.

However, existing approaches for generating synthetic evaluation data are largely heuristic, relying on intuitive, trial-and-error engineering decisions, such as prompt design \citep{wang2023self}, agent setup \citep{li2025eliciting}, and selectively reporting successful outcomes as ``best practices and lessons'' \citep{liubest}. It is important to systematize this process, and that the generated evaluation data achieves \emph{evaluation objectives} \citep{liautobencher}, for example, generating \emph{difficult} evaluation data that exposes weaknesses, or better \emph{separating} a pool of models to offer statistically significant signals about their strengths and weaknesses. While heuristic-based approaches could achieve these goals, they hardly generalize to new model capabilities and evaluation domains without another round of manual trial-and-error. This calls for a general, automated approach to synthetic evaluation data generation with well-defined evaluation objectives.

In this work, \emph{we take an optimization view towards evaluation data}. We propose \ourmethod{}, optimizing a swarm of data generator models towards quantitative objectives with swarm intelligence. Particle swarm optimization (PSO) \citep{kennedy1995particle} is an algorithm that optimizes a swarm of continuous representations for a utility function, guided by particle- and swarm-level utility signals (\Sref{sec:methodology}). Guided by its initial success in multi-LLM collaboration \citep{feng2025heterogeneous, feng2024model}, we employ PSO to optimize data-generator LMs. Specifically, given a static seed dataset, we train an initial swarm of data generators with self-instruct \citep{wang2023self}. For each iteration, we generate evaluation instances with each data generator, evaluate test taker models on them, calculate the evaluation objective metrics (e.g., \emph{difficult}: one minus the max performance of test takers), and employ PSO to optimize the data generator swarm guided by the metrics. This iterative process produces optimized data generators and new evaluation data characterized by specific evaluation objectives, and we propose a suite of five objectives---\emph{difficult, separate, novel, consistent, personalized}---as evaluation data characteristics to optimize towards.

We also propose \textsc{Adversarial Swarms}, an extended algorithm in which a data generator swarm and a test taker swarm co-evolve in an adversarial setting with two steps. In \emph{data-step}, the data swarm takes a PSO optimization step as previously described with the current snapshot of test takers, aiming to generate harder data; In \emph{model-step}, the test taker swarm takes a PSO optimization step guided by their performance on this iteration's generated data. By alternating between the data and model steps, \textsc{Adversarial Swarms} jointly enhances both synthetic data generation and model capabilities.





Extensive experiments with four evaluation domains and eight baselines demonstrate that \ourmethod{} advances the five evaluation objectives, outperforming baselines across four data domains. \textsc{Adversarial Swarms} further improves model performance on a held-out set, outperforming fine-tuning with either static or generated data by 8.2\% on average. Empirical analyses reveal that \ourmethod{} generalizes to unseen test takers, works for tasks with verifiable ground truths, and successfully optimizes compositions of evaluation objectives. In sum, \ourmethod{} uniquely offers an optimization angle to quantitatively measure and improve synthetic evaluation data---enabling the creation of harder dynamic benchmarks, supporting the evaluation of emerging model capabilities, and reducing risks of data contamination and model exposure.

\begin{figure}[t]
    \centering
    \vspace*{-20pt}
    \includegraphics[width=1.0\linewidth]{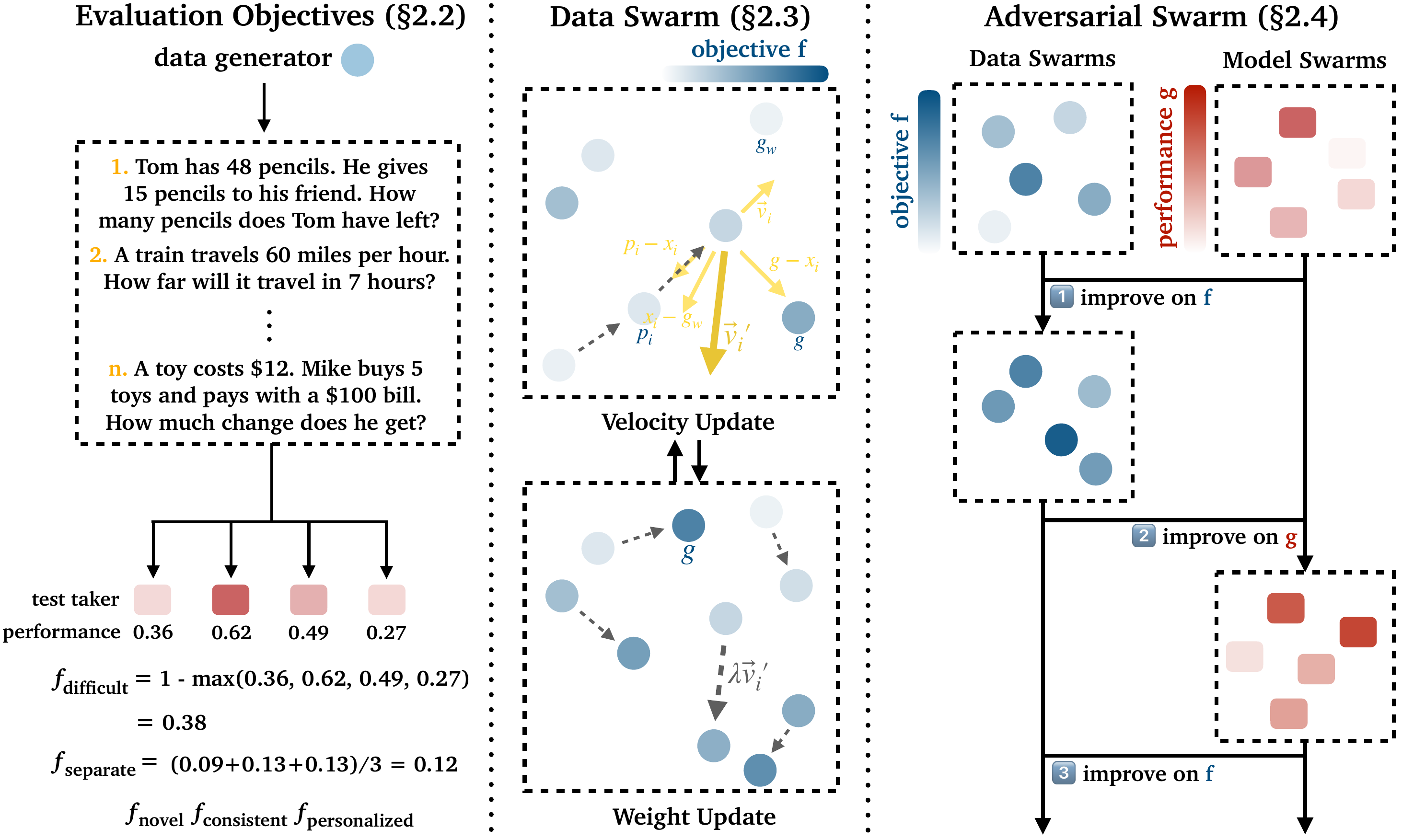}
    \caption{Overview of \ourmethod{}. (Left) Given a data generator, we sample evaluation problems, evaluate test taker LLMs on them, and calculate objectives. (\Sref{subsec:eval_objectives}) (Center) We optimize multiple data generators with particle swarm optimization to adapt them to evaluation objectives. Specifically, models are impacted by the personal/global best/worst checkpoints to update velocity and explore model weight space for optimization. (Right) In \textsc{Adversarial Swarms}, data generators and test takers co-evolve adversarially to produce better data and models simultaneously.}
    \vspace*{-10pt}
    \label{fig:overview}
\end{figure}

\vspace*{-10pt}
\section{Methodology}
\label{sec:methodology}
\vspace*{-5pt}

We propose \ourmethod{}, an algorithm to optimize synthetic evaluation data generation for diverse evaluation objectives (e.g., generating more difficult evaluation examples). Given seed data $\mathcal{D}$ and a pool of test taker models $\{\mathbf{m}_i\}_{i=1}^n$, we train an initial swarm of data generators (\Sref{subsec:data_generators}), define a suite of evaluation objectives (\Sref{subsec:eval_objectives}), and optimize the data generators with particle swarm intelligence (PSO) towards these objectives (\Sref{subsec:data_swarms}). \ourmethod{} transcends the heuristics-driven status quo in synthetic data generation by introducing an optimization-based framework that enables the creation of challenging evaluation data, consistent evaluation, and novel insights into model behavior.

\subsection{Training Initial Data Generators}
\label{subsec:data_generators}

While existing datasets are often offered as a monolithic resource, there is inherent diversity within any dataset \citep{moayeri2024unearthing}: different instances test different variations and mixtures of model capabilities, forming clusters and taxonomies of sub-evaluations \citep{zeng2025evaltree}. Given seed data $\mathcal{D}$ in an existing dataset, we propose to train a swarm of data generators reflecting different sub-evaluations within $\mathcal{D}$ through clustering. By having a pool of multiple data generators, PSO could enable their collaborative search in the model weight space to optimize synthetic data generation (\Sref{subsec:data_swarms}).

Concretely, we first cluster $\mathcal{D}$ into $N$ subsets $(\mathcal{D}_1, \cdots, \mathcal{D}_N) = \mathrm{cluster}(\mathcal{D}, N)$ with K-means over average-pooled query embeddings. For each $\mathcal{D}_i$, we employ Self-Instruct \citep{wang2023self} to train a data generator $\mathbf{x}_i$ to model this distinct cluster of queries: specifically, we randomly sample $(\mathbf{d}_1,\cdots,\mathbf{d}_{2k}) \sim \mathcal{D}_i$ and employ the following format: \emph{``You are an expert in generating synthetic evaluation data, specifically about <domain>. You are given a set of k examples. Please follow the pattern and generate k more examples. Examples: $\mathbf{d}_1, \cdots, \mathbf{d}_k, \underline{\mathbf{d}_{k+1}, \cdots, \mathbf{d}_{2k}}$''}. We then fine-tune an off-the-shelf model with a pool of these prompts into data generator $\mathbf{x}_i$, where $\underline{\mathbf{d}_{k+1}, \cdots, \mathbf{d}_{2k}}$ is the model output for supervised fine-tuning. At inference time, we randomly sample $k$ examples $(\mathbf{d}'_1,\cdots,\mathbf{d}'_{k}) \sim \mathcal{D}_i$ and expect the model to generate $k$ new examples $(\mathbf{d}'_{k+1},\cdots,\mathbf{d}'_{2k}) \sim \mathbf{x}_i(\mathbf{d}'_1,\cdots,\mathbf{d}'_{k})$ for evaluation.

\subsection{Evaluation Objectives}
\label{subsec:eval_objectives}

While existing evaluation efforts have focused on the conceptual and intuitive novelty of datasets, recent research begins to \emph{quantify} the quality and novelty of evaluation data \citep{liautobencher}. Given a data generator or data generation algorithm $\mathbf{x}$ and a pool of test taker models $\{\mathbf{m}_i\}_{i=1}^n$, we generate synthetic data $\mathcal{D}_{\textit{gen}} \sim \mathbf{x}$ and define a suite of the following five \emph{evaluation objectives} $f(\mathbf{x} \mid \{\mathbf{m}_i\}_{i=1}^n)$, which we later seek to advance by optimizing data generators $\{\mathbf{x}_i\}_{i=1}^N$ with swarm intelligence. (\Sref{subsec:data_swarms})

\vspace*{-2mm}
\paragraph{Difficult} Static datasets are often saturated for new state-of-the-art LLMs \citep{open-llm-leaderboard-v2}, thus it has always been important to design difficult and challenging evaluation data. We define the \emph{difficult} objective, aiming to lower performance for the pool of test taker models and expose their weaknesses:

\vspace*{-10pt}
\begin{align*}
    f_{\textit{difficult}}(\mathbf{x} \mid \{\mathbf{m}_i\}_{i=1}^n) = 1 - \mathrm{max}_{i=1}^n \ \mathrm{performance}(\mathbf{m}_i \mid \mathcal{D}_{\textit{gen}})
\end{align*}

\vspace*{-2mm}
\paragraph{Consistent} For a data generator or data generation algorithm $\mathbf{x}$, we could repeatedly sample from it to obtain different generated datasets $(\mathcal{D}_{\textit{gen},1},\cdots,\mathcal{D}_{\textit{gen},k}) \sim \mathbf{x}$. The generated datasets should be robust and \emph{consistent}, i.e.,  model performance on any $\mathcal{D}_{\textit{gen},i}$ should not greatly fluctuate. We calculate the standard deviation $\mathrm{std}$ and define the \emph{consistent} objective as:

\vspace*{-10pt}
\begin{align*}
    f_{\textit{consistent}}(\mathbf{x} \mid \{\mathbf{m}_i\}_{i=1}^n) = 1 - \frac{1}{n} \sum_{i=1}^n \mathrm{std}\big(\{\mathrm{performance}(\mathbf{m}_i) \mid \mathcal{D}_{\textit{gen}, j}\}_{j=1}^k\big)
\end{align*}

\vspace*{-2mm}
\paragraph{Separate} Existing static datasets often result in close performance across models, struggling to distinguish state-of-the-art models and offer insights about their relative strength. We define the \emph{separate} objective, seeking to widen the performance gap between models with generated evaluation data: we first sort $\{\mathrm{performance}(\mathbf{m}_i \mid \mathcal{D}_\textit{gen})\}_{i=1}^n$ in increasing order. $f_{\textit{separate}}$ is then defined as:

\vspace*{-10pt}
\begin{align*}
    f_{\textit{separate}}(\mathbf{x} \mid \{\mathbf{m}_i\}_{i=1}^n) = \frac{1}{n-1} \sum_{i=1}^{n-1} \big( \mathrm{performance}(\mathbf{m}_{i+1} \mid \mathcal{D}_\textit{gen}) - \mathrm{performance}(\mathbf{m}_{i} \mid \mathcal{D}_\textit{gen}) \big)
\end{align*}

\vspace*{-2mm}
\paragraph{Novel} By generating new synthetic evaluation data $\mathcal{D}_\textit{gen}$, we seek to reveal novel insights about model performance different from what is offered by existing data $\mathcal{D}$. We define the \emph{novel} objective as the distance between two performance arrays:

\vspace*{-10pt}
\begin{align*}
    f_{\textit{novel}}(\mathbf{x} \mid \{\mathbf{m}_i\}_{i=1}^n) = \mathrm{distance}(\{\mathrm{performance}(\mathbf{m}_i \mid \mathcal{D}_\textit{gen})\}_{i=1}^n, \{\mathrm{performance}(\mathbf{m}_i \mid \mathcal{D})\}_{i=1}^n)
\end{align*}
where the $\mathrm{distance}$ function could be inverse ranked correlation \citep{liautobencher}, KL divergence, and more distance-based metrics.

\vspace*{-2mm}
\paragraph{Personalized} For user-facing tasks such as instruction following, we should generate instructions that resemble what real-world users might actually ask about. We employ WildChat \citep{zhaowildchat} as a real-world user instruction set $\mathcal{D}_{\textit{user}}$. For generated instruction $\mathbf{d}_{\textit{gen}} \sim \mathcal{D}_{\textit{gen}}$, we employ embedding similarity in retrieval systems to quantify its relevance to user queries $\mathrm{r}(\mathbf{d}_{\textit{gen}}, \mathcal{D}_{\textit{user}}) = \frac{1}{k} \sum_{i=1}^k \mathrm{sim}(\mathbf{d}_{\textit{gen}}, \mathbf{d}_{\textit{user},i})$, the average embedding similarity between the generated instruction and the top-k similar instructions in $\mathcal{D}_{\textit{user}}$. We define the \emph{personalized} objective by averaging over $\mathcal{D}_{\textit{gen}}$:

\vspace*{-10pt}
\begin{align*}
    f_{\textit{personalized}}(\mathbf{x} \mid \{\mathbf{m}_i\}_{i=1}^n) = \sum_{i=1}^{|\mathcal{D}_{\textit{gen}}|} \mathrm{r}(\mathbf{d}_{\textit{gen}, i}, \mathcal{D}_{\textit{user}})
\end{align*}

By defining a suite of five evaluation objectives $f(\mathbf{x} \mid \{\mathbf{m}_i\}_{i=1}^n)$, we comprehensively measure the quality of generated evaluation data and seek to improve data generators $\{\mathbf{x}_i\}_{i=1}^N$ towards these objectives with swarm intelligence. (\Sref{subsec:data_swarms})

\subsection{Data Swarms}
\label{subsec:data_swarms}

\begin{algorithm*}[t]
\SetAlgoLined
\small
\caption{Data Swarms}
\label{alg:data_swarms}
\KwIn{Initial evaluation data $\mathcal{D}$, test taker models $\{\mathbf{m}_i\}_{i=1}^n$, evaluation objective $f(\mathbf{x} \mid \{\mathbf{m}_i\}_{i=1}^n) \rightarrow \mathcal{R}$ where $\mathbf{x}$ is a data generator (\Sref{subsec:eval_objectives}). Hyperparameters: step length $\lambda$, inertia $\phi_v$, cognitive coefficient $\phi_p$, social coefficient $\phi_g$, repel coefficient $\phi_w$, patience $c$, max iteration $\mathcal{K}$.}
train a swarm of initial data generators $\{\mathbf{x}_i\}_{i=1}^N$ on $\mathcal{D}$ (\Sref{subsec:data_generators}) \;
initialize global best checkpoint $\mathbf{g} \gets \varnothing$, global worst checkpoint $\mathbf{g}_w \gets \varnothing$ \;
\For{$i = 1$ \KwTo $N$}{
initialize personal best $\mathbf{p}_i \gets \mathbf{x}_i$, velocity $\mathbf{v}_i \gets \mathbf{0}$ \;
\textbf{if} $f(\mathbf{x}_i \mid \{\mathbf{m}_i\}_{i=1}^n) > f(\mathbf{g} \mid \{\mathbf{m}_i\}_{i=1}^n)$, $\mathbf{g} \leftarrow \mathbf{x}_i$; \textbf{if} $f(\mathbf{x}_i \mid \{\mathbf{m}_i\}_{i=1}^n) < f(\mathbf{g}_w \mid \{\mathbf{m}_i\}_{i=1}^n)$, $\mathbf{g}_w \leftarrow \mathbf{x}_i$
}
\For{$k = 1$ \KwTo $\mathcal{K}$}{
\lIf{$\mathbf{g} \ \mathrm{did \ not \ change \ in \ the \ last} \ c \ \mathrm{iterations}$}{break}
\For{$i = 1$ \KwTo $N$ $\mathrm{parallel}$}{
randomness factors $r_v, r_p, r_g, r_w \sim \mathcal{U}(0,1)$ \;
velocity $\mathbf{v}_i \gets \frac{1}{\mathcal{C}}[r_v\phi_v\mathbf{v}_i + r_p\phi_p(\mathbf{p}_i - \mathbf{x}_i) + r_g\phi_g(\mathbf{g} - \mathbf{x}_i) - r_w\phi_w(\mathbf{g}_w - \mathbf{x}_i)]$, where normalization term $\mathcal{C} = r_v\phi_v + r_p\phi_p + r_g\phi_g + r_w\phi_w$ \;
model weights $\mathbf{x}_i \gets \mathbf{x}_i + \lambda \mathbf{v}_i$ \;
\textbf{if} $f(\mathbf{x}_i \mid \{\mathbf{m}_i\}_{i=1}^n) > f(\mathbf{g} \mid \{\mathbf{m}_i\}_{i=1}^n)$, $\mathbf{g} \leftarrow \mathbf{x}_i$; \textbf{if} $f(\mathbf{x}_i \mid \{\mathbf{m}_i\}_{i=1}^n) < f(\mathbf{g}_w \mid \{\mathbf{m}_i\}_{i=1}^n)$, $\mathbf{g}_w \leftarrow \mathbf{x}_i$; \textbf{if} $f(\mathbf{x}_i \mid \{\mathbf{m}_i\}_{i=1}^n) > f(\mathbf{p}_i \mid \{\mathbf{m}_i\}_{i=1}^n)$, $\mathbf{p}_i \leftarrow \mathbf{x}_i$ \;
}
}
\Return $\mathbf{g}$
\end{algorithm*}

We propose \ourmethod{}, a collaborative search algorithm to optimize data generators $\{\mathbf{x}_i\}_{i=1}^N$ with particle swarm optimization (PSO) \citep{kennedy1995particle}. Simply put, we iteratively learn a weighted combination of data generator LLMs using the PSO algorithm. Specifically, the model weights of each $\mathbf{x}_i$ represent its \emph{location} in the space of model weights. Evaluation objectives $f(\mathbf{x}_i \mid \{\mathbf{m}_i\}_{i=1}^n)$ are the utility function to be optimized in the model weight space. We track several variables for the data swarm:

\begin{itemize}[leftmargin=*]
    \item swarm-level: \emph{global best} $\mathbf{g}$, the best-found model weights across all of $\{\mathbf{x}_i\}_{i=1}^N$'s search trajectories in the weight space; and conversely \emph{global worst} $\mathbf{g}_w$.
    \item individual-level: \emph{personal best} $\mathbf{p}_i$, the best-found model weights in $\mathbf{x}_i$'s search trajectory; \emph{velocity} $\mathbf{v}_i$, the direction in the weight space that the model move next, initialized as $\mathbf{0}$.
\end{itemize}

In each step, we update the velocity of models guided by the individual- and swarm-level signals:

\begin{align*}
    \mathbf{v}_i \gets \frac{1}{\mathcal{C}}\big[r_v\phi_v\mathbf{v}_i + r_p\phi_p(\mathbf{p}_i - \mathbf{x}_i) + r_g\phi_g(\mathbf{g} - \mathbf{x}_i) - r_w\phi_w(\mathbf{g}_w - \mathbf{x}_i)\big]
\end{align*}
where the inertia term $r_v\phi_v\mathbf{v}_i$ maintains part of the model's current velocity to chat an independent search path; $r_p\phi_p(\mathbf{p}_i - \mathbf{x}_i)$, $r_g\phi_g(\mathbf{g} - \mathbf{x}_i)$, $- r_w\phi_w(\mathbf{g}_w - \mathbf{x}_i)$ indicates that the model is drawn by personal/global best and repelled by global worst to further explore/escape from these weight areas due to continuity in the model weight neighborhood \citep{eilertsen2020classifying}; $r_v, r_p, r_g, r_w \sim \mathcal{U}(0,1)$ are randomness factors to enable search stochasticity to explore out of local minima; $\phi_v, \phi_p, \phi_g, \phi_w$ are hyperparameters and $\mathcal{C} = r_v\phi_v + r_p\phi_p + r_g\phi_g + r_w\phi_w$ is a normalization term.

Each model then takes a step towards the updated velocty direction $\mathbf{x}_i \gets \mathbf{x}_i + \lambda \mathbf{v}_i$, evaluate on $f(\mathbf{x}_i \mid \{\mathbf{m}_i\}_{i=1}^n)$, and update personal/global best/worst checkpoints if they have changed. By iteractively updating the velocity and model weights, the swarm of data generators collaboratively search in the weight space to improve evaluation objectives. In the end, the global best checkpoint $\mathbf{g}$, representing the best-found data generator in the search process, is retained and employed to generate synthetic evaluation data. We summarize \ourmethod{} in Algorithm \ref{alg:data_swarms}.

\subsection{Adversarial Swarms}
\label{subsec:dual_swarms}

\begin{algorithm*}[t]
\SetAlgoLined
\small
\caption{Adversarial Swarms}
\label{alg:dual_swarms}
\KwIn{Initial evaluation data $\mathcal{D}$, test taker models $\{\mathbf{m}_i\}_{i=1}^n$, evaluation objective $f(\mathbf{x} \mid \{\mathbf{m}_i\}_{i=1}^n) \rightarrow \mathcal{R}$ where $\mathbf{x}$ is a data generator (\Sref{subsec:eval_objectives}). Hyperparameters: max iteration $\mathcal{K}$, patience $c$.} \vspace{5pt}
train a swarm of initial data generators $\{\mathbf{x}_i\}_{i=1}^N$ on $\mathcal{D}$ (\Sref{subsec:data_generators}) (lines 2-6, Alg.\ref{alg:data_swarms}) \;
initialize the data swarm $\{\mathbf{x}_i\}_{i=1}^N$ and model swarm $\{\mathbf{m}_i\}_{i=1}^n$ \;
\For{$k = 1$ \KwTo $\mathcal{K}$}{
\lIf{$\mathrm{the \ two \ global \ bests \ } \mathbf{g}_{\textit{data}} \mathrm{\ and \ } \mathbf{g}_{\textit{model}} \ \mathrm{did \ not \ change \ in \ the \ last} \ c \ \mathrm{iterations}$}{break}
$\{\mathbf{x}_i\}_{i=1}^N \leftarrow \mathrm{PSO \mhyphen step}\footnotemark[2](\{\mathbf{x}_i\}_{i=1}^N, \{f(\mathbf{x}_i \mid \{\mathbf{m}_j\}_{j=1}^n)\}_{i=1}^N)$ (lines 10-13, Alg.\ref{alg:data_swarms}) \;
generate joint dataset $\mathcal{D}_{\textit{joint}} = \bigcup_{i=1}^N \mathrm{sample}(\mathbf{x}_i)$ \;
$\{\mathbf{m}_i\}_{i=1}^n \leftarrow \mathrm{PSO \mhyphen step}\footnotemark[2](\{\mathbf{m}_i\}_{i=1}^n, \{\mathrm{performance}(\mathbf{m}_i \mid \mathcal{D}_{\textit{joint}})\}_{i=1}^n)$ \;
}
\Return global bests $\mathbf{g}_{\textit{data}}$ and $\mathbf{g}_{\textit{model}}$
\end{algorithm*}\footnotetext[2]{$\mathrm{PSO \mhyphen step}(\{\mathbf{x}_i\}_{i=1}^N, \{v_i\}_{i=1}^N$) indicates taking one PSO step to update the model weights of $\{\mathbf{x}_i\}_{i=1}^N$ with their scalar scores $\{v_i\}_{i=1}^N$, lines 10 to 13 in Algorithm \ref{alg:data_swarms}.}

In \ourmethod{}, the data generators evolve while the pool of test taker models $\{\mathbf{m}_i\}_{i=1}^n$ stay frozen. If test takers could co-evolve to learn from the generated data, this would lead to stronger models and place a higher demand for data generators to further improve. To this end, we extend \ourmethod{} into \textsc{Adversarial Swarms}, where the \emph{data} generator \emph{swarm} and test taker \emph{model swarm} both evolve in an adversarial way. We first initialize the velocity and personal/global checkpoints for both the data swarm $\{\mathbf{x}_i\}_{i=1}^N$ and the model swarm $\{\mathbf{m}_i\}_{i=1}^n$. (lines 2-6, Algorithm \ref{alg:data_swarms}) \textsc{Adversarial Swarms} then alternate between two steps:

In \emph{data-step}, the test taker models $\{\mathbf{m}_i\}_{i=1}^n$ are fixed while the data generators $\{\mathbf{x}_i\}_{i=1}^N$ take an particle swarm optimization step, guided by the evaluation objective metrics:

\begin{align*}
    \{\mathbf{x}_i\}_{i=1}^N \leftarrow \mathrm{PSO \mhyphen step}(\{\mathbf{x}_i\}_{i=1}^N, \{f(\mathbf{x}_i \mid \{\mathbf{m}_j\}_{j=1}^n)\}_{i=1}^N)
\end{align*}
where $\mathrm{PSO \mhyphen step}$ denotes updating the velocity and model weights of $\{\mathbf{x}_i\}_{i=1}^N$ guided by their achieved evaluation objective metric $\{f(\mathbf{x}_i \mid \{\mathbf{m}_j\}_{j=1}^n)\}_{i=1}^N$. (lines 10-13, Algorithm \ref{alg:data_swarms})

In \emph{model-step}, the data generators are fixed and we sample a joint datset from them $\mathcal{D}_{\textit{joint}} = \bigcup_{i=1}^N \mathrm{sample}(\mathbf{x}_i)$. The swarm of test taker models are then evaluated on $\mathcal{D}_{\textit{joint}}$, and their performance becomes a signal for PSO update:

\begin{align*}
    \{\mathbf{m}_i\}_{i=1}^n \leftarrow \mathrm{PSO \mhyphen step}(\{\mathbf{m}_i\}_{i=1}^n, \{\mathrm{performance}(\mathbf{m}_i \mid \mathcal{D}_{\textit{joint}})\}_{i=1}^n)
\end{align*}

We by default employ the difficult evaluation objective $f_{\textit{difficult}}$: in this setting, data generators are optimized for harder data (lowering model performance) while test taker models are optimized to improve on such data (improving model performance), inducing an adversarial setup where the data generators and test takers compete and co-evolve for stronger models and evaluation. In the end, the global best checkpoints $\mathbf{g}_{\textit{data}}$ and $\mathbf{g}_{\textit{model}}$ are retained as a harder data generator and a stronger model. We empirically observe that having a first-in-last-out sliding \emph{window} of generated data, instead of completely swapping out with newly generated data in $\mathcal{D}_{\textit{joint}}$, yields better performance and we follow this setting. We summarize \ourmethod{} in Algorithm \ref{alg:dual_swarms}.

\section{Experiment Settings}
\label{sec:experiment_settings}
\vspace*{-5pt}

\paragraph{Models and Implementation} We employ \textsc{Gemma-9B} (\emph{google/gemma-2-9b-it}) \citep{team2024gemma} as the base model for data generator fine-tuning (\Sref{subsec:data_generators}), sample 10k in-context learning instances with $k=5$, and fine-tune for 5 epochs with $2e-4$ starting learning rate and batch size $32$ by default. We employ a pool of 4 test taker models by fine-tuning \textsc{Gemma-9B} on 4 SFT domains in Tulu-v2 \citep{ivison2023camels} (\emph{cot}, \emph{lima} \citep{zhou2023lima}, \emph{oasst1} \citep{kopf2023openassistant}, and \emph{science}) to induce heterogeneous skills, while investigating three other test taker settings in Section \ref{sec:analysis}. We employ KL-divergence for $\mathrm{distance}$ in the novel objective and employ WildChat \citep{zhaowildchat} as the real-world query repository in the personalized objective. For \ourmethod{} searches, we employ $c = 5$, $\mathcal{K} = 30$, generate 200 evaluation instances for each iteration and each data generator, while running grid search over other hyperparameters and report the best-found data generator based on evaluation objective $f$, details in Appendix \ref{sec:details_appendix}.
\vspace*{-5pt}

\paragraph{Baselines} For \ourmethod{}, we compare with eight baselines of data and data generation methods: the existing \textsc{dev set}, the existing \textsc{held-out set}, initial data generators in Sec. \ref{subsec:data_generators} (\textsc{init. gen}), \textsc{model soups} \citep{wortsman2022model} over initial data generators, \textsc{self-instruct} \citep{wang2023self}, \textsc{autobencher} \citep{liautobencher}, \textsc{task elicitation} \citep{brown2025adaptively}, and \textsc{prompt breeder} \citep{fernandopromptbreeder}. The first four are objective-agnostic approaches while the latter four are guided by the same evaluation objectives in Sec. \ref{subsec:eval_objectives} as \ourmethod{}.
\vspace*{-5pt}

\paragraph{Data and Evaluation} We employ four datasets as the starter data $\mathcal{D}$: instruction following with \textsc{alpaca} \citep{dubois2024alpacafarm}, math with \textsc{gsm8k} \citep{cobbe2021training}, factuality with \textsc{truthfulqa} \citep{lin2022truthfulqa}, and knowledge capabilities with \textsc{wikidyk} \citep{zhang2025wikidyk}. We employ LLM-as-a-judge \citep{zheng2023judging} with \textsc{gemini-1.5-flash} to score answers in 1-10 and normalize it to 0-1, verify its reliability in Appendix \ref{sec:details_appendix}, while exploring tasks with natural ground truths in Section \ref{sec:analysis}. We by default sample 2k and 1k examples for the dev set and the held-out set unseen at optimization time.
\vspace*{-5pt}

\begin{table}[t]
\centering
\scriptsize
\setlength{\tabcolsep}{2pt}
\renewcommand{\arraystretch}{1.1}
\vspace*{-30pt}
\resizebox{0.9\linewidth}{!}{
\begin{tabular}{lcccccccc}\toprule[1.5pt]
&\multicolumn{4}{c}{Alpaca} &\multicolumn{4}{c}{GSM8k} \\\cmidrule{2-5} \cmidrule{6-9}
&Difficult &Separate &Novel &Consistent\footnotemark[3] &Difficult &Separate &Novel &Consistent \\\midrule
\textcolor{NavyBlue}{\textsc{Dev Set}} &.3201 &.1016 &.0000 &/ &.0711 &.0487 &.0000 &/ \\
\textcolor{NavyBlue}{\textsc{Held-Out}} &.3157 &\underline{.1049} &.0001 &/ &.0850 &.0506 &.0003 &/ \\
\textcolor{NavyBlue}{\textsc{Init. Gen}} &.2955 {\tiny (.0262)} &.0790 {\tiny (.0124)} &.0016 {\tiny (.0009)} &\underline{.9849} {\tiny (.0039)} &.1380 {\tiny (.0233)} &.0459 {\tiny (.0076)} &.0005 {\tiny (.0002)} &\underline{.9869} {\tiny (.0020)} \\
\textcolor{NavyBlue}{\textsc{Model Soups}} &.2380 {\tiny (.0194)} &.0968 {\tiny (.0020)} &.0037 {\tiny (.0024)} &.9785 {\tiny (.0057)} &.0892 {\tiny (.0146)} &.0432 {\tiny (.0093)} &.0004 {\tiny (.0002)} &.9776 {\tiny (.0071)} \\ \midrule
\textcolor{Orange}{\textsc{Self-Inst}} &.2087 {\tiny (.0050)} &.0822 {\tiny (.0095)} &.0077 {\tiny (.0102)} &.9415 {\tiny (.0231)} &.0499 {\tiny (.0135)} &\underline{.0571} {\tiny (.0137)} &\underline{.0033} {\tiny (.0016)} &.9327 {\tiny (.0211)} \\
\textcolor{Orange}{\textsc{Autobench}} &.2347 {\tiny (.0086)} &.0801 {\tiny (.0070)} &.0070 {\tiny (.0047)} &.9736 {\tiny (.0071)} &.1268 {\tiny (.0114)} &.0569 {\tiny (.0049)} &.0010 {\tiny (.0001)} &.9753 {\tiny (.0080)} \\
\textcolor{Orange}{\textsc{Task Elicit}} &\underline{.3315} {\tiny (.0148)} &.0877 {\tiny (.0048)} &.0074 {\tiny (.0012)} &.9790 {\tiny (.0141)} &.0980 {\tiny (.0171)} &.0376 {\tiny (.0091)} &.0024 {\tiny (.0009)} &.9795 {\tiny (.0371)} \\
\textcolor{Orange}{\textsc{Prompt Breed}} &.2423 {\tiny (.0189)} &.0959 {\tiny (.0112)} &\underline{.0092} {\tiny (.0065)} &.9813 {\tiny (.0046)} &\underline{.1713} {\tiny (.0254)} &.0554 {\tiny (.0034)} &.0018 {\tiny (.0003)} &.9779 {\tiny (.0071)} \\ \midrule
\textcolor{OliveGreen}{\ourmethod{}} &\textbf{.3868} {\tiny (.0218)} &\textbf{.1177} {\tiny (.0009)} &\textbf{.0113} {\tiny (.0014)} &\textbf{.9925} {\tiny (.0005)} &\textbf{.2865} {\tiny (.0119)} &\textbf{.0767} {\tiny (.0029)} &\textbf{.0226} {\tiny (.0059)} &\textbf{.9934} {\tiny (.0013)} \\ \midrule
&\multicolumn{4}{c}{TruthfulQA} &\multicolumn{4}{c}{WikiDYK} \\ \cmidrule{2-5} \cmidrule{6-9}
&Difficult &Separate &Novel &Consistent &Difficult &Separate &Novel &Consistent \\ \midrule
\textcolor{NavyBlue}{\textsc{Dev Set}} &.2464 &.1051 &.0000 &/ &\underline{.3766} &.0313 &.0000 &/ \\
\textcolor{NavyBlue}{\textsc{Held-Out}} &.2557 &.1042 &.0022 &/ &.3489 &.0316 &.0005 &/ \\
\textcolor{NavyBlue}{\textsc{Init. Gen}} &.1512 {\tiny (.0217)} &.0932 {\tiny (.0140)} &.0011 {\tiny (.0008)} &\underline{.9893} {\tiny (.0049)} &.3643 {\tiny (.0345)} &.0376 {\tiny (.0094)} &.0009 {\tiny (.0002)}&\underline{.9839} {\tiny (.0010)} \\
\textcolor{NavyBlue}{\textsc{Model Soups}} &.1987 {\tiny (.0125)} &.0766 {\tiny (.0126)} &.0022 {\tiny (.0015)} &.9800 {\tiny (.0076)} &.3699 {\tiny (.0028)} &.0293 {\tiny (.0020)} &.0010 {\tiny (.0007)} &.9810 {\tiny (.0080)} \\ \midrule
\textcolor{Orange}{\textsc{Self-Inst}} &.1482 {\tiny (.0260)} &.0884 {\tiny (.0168)} &.0028 {\tiny (.0016)} &.9595 {\tiny (.0169)} &.2038 {\tiny (.0307)} &.0373 {\tiny (.0165)} &.0037 {\tiny (.0028)} &.9029 {\tiny (.0108)} \\
\textcolor{Orange}{\textsc{Autobench}} &.2068 {\tiny (.0058)} &.0846 {\tiny (.0197)} &.0019 {\tiny (.0009)} &.9727 {\tiny (.0043)} &.1527 {\tiny (.0149)} &.0218 {\tiny (.0023)} &.0015 {\tiny (.0005)} &.9683 {\tiny (.0055)} \\
\textcolor{Orange}{\textsc{Task Elicit}} &\underline{.2732} {\tiny (.0110)} &.0980 {\tiny (.0100)} &\underline{.0064} {\tiny (.0005)} &.9618 {\tiny (.0095)} &.3242 {\tiny (.0068)} &\underline{.0423} {\tiny (.0083)} &\textbf{.0068} {\tiny (.0008)} &.9771 {\tiny (.0063)} \\
\textcolor{Orange}{\textsc{Prompt Breed}} &.2235 {\tiny (.0184)} &\underline{.1233} {\tiny (.0103)} &.0045 {\tiny (.0013)}&.9795 {\tiny (.0052)} &.3328 {\tiny (.0403)} &.0310 {\tiny (.0057)} &.0031 {\tiny (.0005)} &.9763 {\tiny (.0040)} \\ \midrule
\textcolor{OliveGreen}{\ourmethod{}} &\textbf{.3593} {\tiny (.0117)} &\textbf{.1535} {\tiny (.0055)} &\textbf{.0461} {\tiny (.0076)} &\textbf{.9932} {\tiny (.0011)} &\textbf{.4662} {\tiny (.0055)} &\textbf{.0614} {\tiny (.0006)} &\underline{.0058} {\tiny (.0004)} &\textbf{.9898} {\tiny (.0005)} \\
\bottomrule[1.5pt]
\end{tabular}
}
\vspace*{5pt}
\caption{Evaluation objective results with different data generation methods. Static dev and held-out sets are not from sampling and do not have consistent metrics. We repeat each data generation methods for five times and report the value and (standard deviation). Best in \textbf{bold} and second-best in \underline{underline}. \textcolor{OliveGreen}{\ourmethod{}} consistently outperforms \textcolor{NavyBlue}{objective-agnostic} and \textcolor{Orange}{objective-guided} approaches and generates more difficult, separating, novel, and consistent evaluation data compared to baselines.}
\vspace*{-20pt}
\label{tab:big}
\end{table}
\footnotetext[3]{The consistent objective is one minus the standard deviation of model performances (typically $<<0.1$), thus the values are high and any improvement in the absolute value have a larger impact than it seems.}

\section{Results}
\label{sec:results}

\vspace*{-2mm}
\paragraph{Data Swarms} We present performance of \ourmethod{} and data generation baselines on four evaluation objectives (\Sref{subsec:eval_objectives}) in Table \ref{tab:big}. \ourmethod{} consistently outperforms all eight baselines in 15 of the 16 dataset and objective settings. \ourmethod{} works especially well in generating math reasoning problems, with 37.2\% higher average improvement than the other three datasets: through qualitative analysis (\Sref{sec:analysis}) we find that \ourmethod{} often generates problems with longer context and are more compositional that pose greater challenges to the test taker models. \textcolor{Orange}{Objective-guided} data generation baselines outperform \textcolor{NavyBlue}{objective-agnostic} approaches by 35.9\% on average, while \ourmethod{} further outperforms them by providing stronger optimization through model collaboration in the weight space and better adapts data generators to evaluation objectives.

\begin{figure}[t]
    \centering
    \vspace*{-25pt}
    \includegraphics[width=1\linewidth]{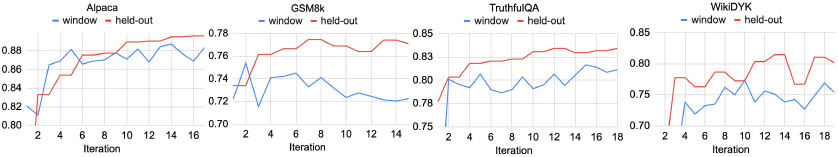}
    \vspace*{-20pt}
    \caption{Performance on the window set and held-out set in \textsc{Adversarial Swarms}. Models fluctuate on the window set while generalizing to the held-out set for consistent improvement.}
    \vspace*{-10pt}
    \label{fig:dual_lines}
\end{figure}

\begin{wrapfigure}{r}{0.4\textwidth}
    \centering
    \vspace*{-10pt}
    \includegraphics[width=0.4\textwidth]{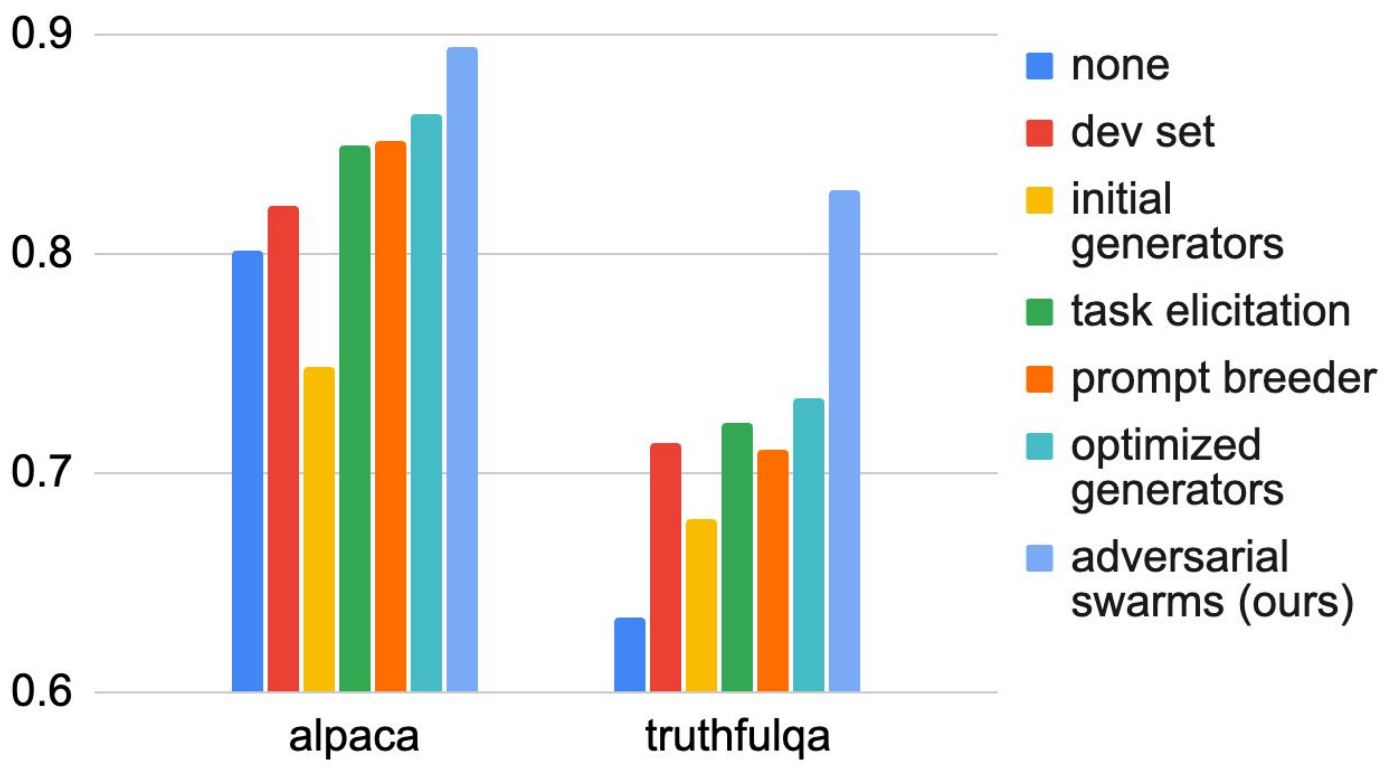}
    \caption{Performance on the test set when fine-tuning on different data settings and using \textsc{Adversarial Swarms}: ours outperforms fine-tuning by offering co-evolution of data generators and test takers in competition.}
    \vspace*{-10pt}
    \label{fig:dual_sft}
\end{wrapfigure}

\vspace*{-2mm}
\paragraph{Adversarial Swarms} Data generators and test taker models compete on the sliding \emph{window} subset of data (\Sref{subsec:dual_swarms}): data generators seek to generate harder questions (improving $f_{\textit{difficult}}$) as the window set, and test takers improve their performance on the generated data in each iteration. We additionally employ a held-out set of data, not visible to either data generators or test takers, to monitor generalization. We present performance of the best test taker on the window and held-out set in Figure \ref{fig:dual_lines}. We observe \emph{fluctuation} of window set performance, indicating that data generators and test takers are competing to lower the objective/performance of each other. However, we see \emph{consistent improvement} on the held-out set, indicating that by adversarially generating harder problems and learning from them, models are learning from the increasingly harder data and obtain generalized skills in the domain.

\begin{wrapfigure}{r}{0.4\textwidth}
    \centering
    \vspace*{-25pt}
    \includegraphics[width=0.4\textwidth]{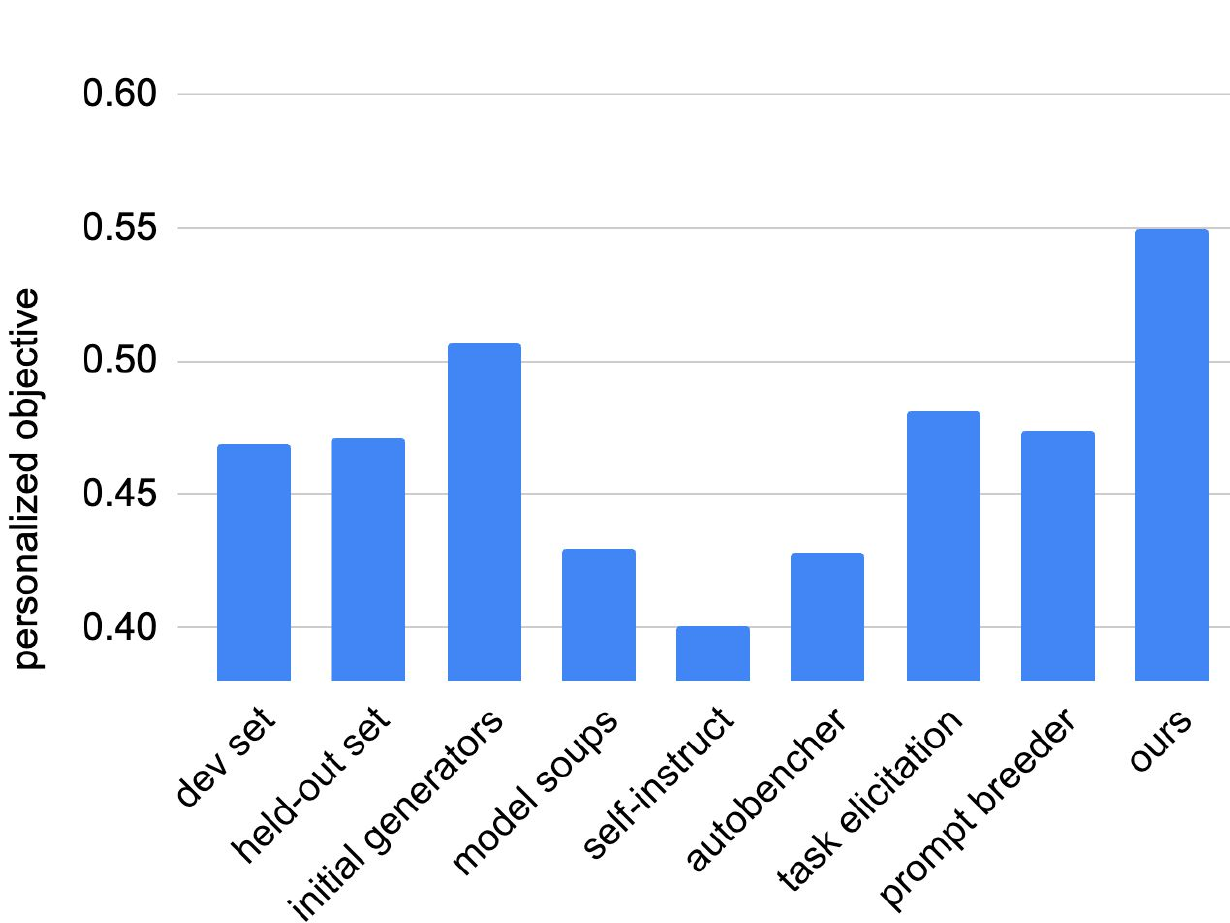}
    \vspace*{-15pt}
    \caption{Results for the personalized objective $f_{\textit{personalized}}$ on Alpaca. \ourmethod{} outperforms baselines by 8.4\%, flexibly adapting to real-world user instructions.}
    \vspace*{-10pt}
    \label{fig:user_centric}
\end{wrapfigure}

We compare the best-found test taker model in \textsc{Adversarial Swarms} with other ways of leveraging synthetic data on a separate test set, specifically by fine-tuning the base model on different generated data: \emph{none}, simply base model performance; fine-tuning on the \emph{dev set} employed by the data generator swarms; fine-tuning on the data from \emph{initial data generators} without particle swarm optimization; fine-tuning on the \emph{optimized data generators} without the test taker swarm; fine-tuning on the data generated by two baselines \emph{task elicitation} and \emph{prompt breeder}. If \textsc{Adversarial Swarms} ends in $n$ iterations, we sample $n / 2 $ times more data from baselines to maintain the same data size (half of the iterations are data-step) for fair comparison. Results in Figure \ref{fig:dual_sft} demonstrate that \textsc{Adversarial Swarms} consistently outperforms these fine-tuning settings: instead of test taker models passively accepting generated data, \textsc{Adversarial Swarms} uniquely enables the co-evolution of data generator and test takers for flexible and interactive learning from synthetic data.

\vspace*{-10pt}
\section{Analysis}
\label{sec:analysis}
\vspace*{-10pt}

\paragraph{Objective: Personalized} In addition to generating difficult, separating, novel, and consistent evaluation data, we posit that the generated queries should be \emph{personalized}, resembling what users might actually ask about in human-AI interactions. We employ Alpaca as the starter dataset, WildChat \citep{zhaowildchat} as the user instruction repository for similarity search, and evaluate \ourmethod{} and baselines on the personalized objective $f_{\textit{personalized}}$ (\Sref{subsec:eval_objectives}). Results in Figure \ref{fig:user_centric} demonstrate that \ourmethod{} improves over baselines by at least 8.4\% on $f_{\textit{user}}$, indicating that \ourmethod{} flexibly adapts to real-world user instructions. Note that while the first four objectives were calculated based on the performance of test taker models, $f_{\textit{user}}$ is \emph{independent} of the model being evaluated. This suggests that \ourmethod{} could both adapt to model-specific and model-agnostic evaluation objectives.

\begin{table}[t]
\centering
\scriptsize
\setlength{\tabcolsep}{3pt}
\renewcommand{\arraystretch}{1.1}
\vspace*{-20pt}
\resizebox{0.9\linewidth}{!}{
\begin{tabular}{lcccccccccccc}\toprule[1.5pt]
&\multicolumn{4}{c}{Setting 1: Qwen Models} &\multicolumn{4}{c}{Setting 2: Open Models} &\multicolumn{4}{c}{Setting 3: Closed Models} \\\cmidrule{2-13}
&Alpaca &GSM8k &Truthful &WikiDYK &Alpaca &GSM8k &Truthful &WikiDYK &Alpaca &GSM8k &Truthful &WikiDYK \\\midrule[0.75pt]
\textcolor{NavyBlue}{\textsc{Dev Set}} &.400 &.106 &.383 &.551 &.105 &.016 &.071 &.184 &.094 &.013 &.077 &.113 \\
\textcolor{NavyBlue}{\textsc{Held-Out Set}} &.381 &.119 &.266 &.533 &.106 &.023 &.070 &.177 &.092 &.008 &.075 &.081 \\
\textcolor{Orange}{\textsc{Task Elicit}} &.444 &.219 &.385 &.520 &.214 &.158 &.077 &.190 &.157 &.215 &.072 &.099 \\
\textcolor{Orange}{\textsc{Prompt Breed}} &.413 &.250 &\bf .446 &.460 &.143 &\bf  .211 &.102 &.205 &.104 &.220 &.037 &.091 \\ \midrule[0.75pt]
\textcolor{OliveGreen}{\ourmethod{}} &\bf .486 &\bf .259 &.414 &\bf .621 &\bf .308 &.193 &\bf .129 &\bf .257 &\bf \bf .234 &\bf .223 &\bf .096 &\bf .130 \\
\bottomrule[1.5pt]
\end{tabular}
}
\vspace*{5pt}
\caption{Evaluating generated data optimized for test takers $\mathcal{M}$ on three settings of unseen test takers $\mathcal{M}'$. \ourmethod{} outperforms baselines in 10 of the 12 settings, discovering challenging and high-quality evaluation data generalizable to models in the wild.}
\vspace*{-20pt}
\label{tab:unseen_takers}
\end{table}

\vspace*{-2mm}
\paragraph{Natural Ground Truths} When generating harder and novel problems, we often don't have high-quality ground truths without human annotation \citep{liautobencher}. LLM-as-a-judge is widely employed as a quick substitution \citep{zheng2023judging}, while for some tasks there exists \emph{natural ground truths}, e.g. problems that could be solved with symbolic systems or algorithms. We employ graph algorithm reasoning, specifically the NLGraph dataset \citep{wang2023can}, as starter data $\mathcal{D}$, use graph algorithms to derive ground truths and evaluate (e.g., Dijkstra's to evaluate LLM shortest path reasoning), and optimize \ourmethod{} for $f_{\textit{difficult}}$ and $f_{\textit{separate}}$. Table \ref{tab:ground_truth} demonstrates that \ourmethod{} also better adapts to evaluation domains with natural ground truths, outperforming baselines by 6.5 and 6.2 points across the two objectives.

\begin{wraptable}{r}{0.4\textwidth}
\centering
\scriptsize
\vspace*{-10pt}
\resizebox{1\linewidth}{!}{
\begin{tabular}{lccc}\toprule[1.5pt]
&Difficult &Separate \\\midrule[0.75pt]
\textcolor{NavyBlue}{\textsc{Dev Set}} &0.524 &0.112 \\
\textcolor{NavyBlue}{\textsc{Held-Out Set}} &0.544 &0.081 \\
\textcolor{Orange}{\textsc{Task Elicit}} &0.574 &0.132 \\
\textcolor{Orange}{\textsc{Prompt Breed}} &0.555 &0.124 \\
\textcolor{OliveGreen}{\ourmethod{}} &\bf 0.639 &\bf 0.194 \\
\bottomrule[1.5pt]
\end{tabular}
}
\caption{Results on generating graph problems where ground truth is naturally available. \ourmethod{} outperforms baselines by 6.35 points on average.}
\vspace*{-20pt}
\label{tab:ground_truth}
\end{wraptable}

\vspace*{-2mm}
\paragraph{Generalizing to Unseen Test Takers} \ourmethod{} assumes access to a pool of test taker models $\mathcal{M} = \{\mathbf{m}_i\}_{i=1}^n$: when new data is generated, we evaluate $\mathcal{M}$ on them to calculate the objectives in Section \ref{subsec:eval_objectives}. In addition to generating problems that are harder/more novel for $\mathcal{M}$, would those quantitative properties generalize to other test takers $\mathcal{M}'$, unseen in the optimization process? We employ the best-found generated data for baselines and \ourmethod{} based on $\mathcal{M}$ (four differently fine-tuned Gemma models, \Sref{sec:experiment_settings}), and calculate their $f_{\textit{difficult}}$ scores with three settings of $\mathcal{M}'$ to test generalization:

\begin{figure}[t]
    \centering
    \includegraphics[width=1\linewidth]{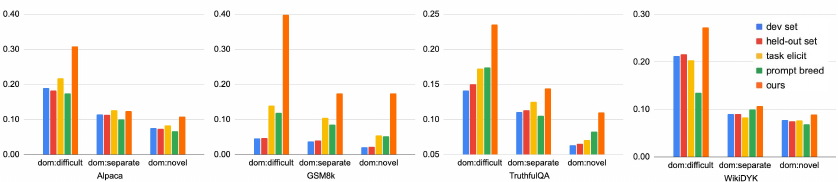}
    \vspace*{-20pt}
    \caption{Evaluating \ourmethod{} and baselines with composite evaluation objectives, weighted averages of difficult, separate, and novel objectives with one being \emph{dominant} and having the largest weight. \ourmethod{} outperforms baselines in ten of the twelve dataset and objective settings.}
    \vspace*{-10pt}
    \label{fig:objective_composition}
\end{figure}

\begin{itemize}[leftmargin=*]
    \item \emph{$\mathcal{M} \rightarrow$ Qwen}: We fine-tune \emph{Qwen2.5-7B-Instruct} on four different SFT mixtures in Tulu-v2 \citep{ivison2023camels} (\Sref{sec:experiment_settings}), resulting in four models with varying expertise as $\mathcal{M}'$.
    \item \emph{$\mathcal{M} \rightarrow$ open}: \emph{gemma-2-9b-it}, \emph{Qwen2.5-7B-Instruct}, \emph{Llama-3.1-8B-Instruct}, and \emph{DeepSeek-R1-Distill-Qwen-7B} off-the-shelf as $\mathcal{M}'$.
    \item \emph{$\mathcal{M} \rightarrow$ closed}: \emph{gpt-4o}, \emph{gpt-4o-mini}, \emph{gemini-1.5-flash-002}, and \emph{gemini-1.5-pro-002} as $\mathcal{M}'$.
\end{itemize}

Table \ref{tab:unseen_takers} demonstrates that \ourmethod{} better generalizes to unseen test takers across datasets and settings, outperforming baselines by 4.6\%, 16.0\%, and 22.4\% on average in the three generalization settings. This suggests that \ourmethod{} successfully pinpoints common model weaknesses to generate more challenging problems, generalizing from target models to models in the wild.

\vspace*{-2mm}
\paragraph{Objective Composition} In addition to optimizing one single objective, we often pursue a mixture of objectives (e.g. evaluation data that are hard and novel at the same time). We investigate whether \ourmethod{} and baselines could optimize objective compositions where one goal is prioritized more than others: $f_{\textit{dom:difficult}} = w_{\textit{h}} f_{\textit{difficult}} + w_{\textit{l}} f_{\textit{separate}} + w_{\textit{l}} f_{\textit{novel}}$, similarly for $f_{\textit{dom:separate}}$ and $f_{\textit{dom:novel}}$. We employ $w_h = 0.6$ and $w_l = 0.2$ and present results in Figure \ref{fig:objective_composition}. Results demonstrate \ourmethod{} consistent improves over baselines by 6.2 points on average, while we observe greater improvement when $f_{\textit{difficult}}$ is the dominant objective, consistent with our findings for Table \ref{tab:big}.

\vspace*{-2mm}
\paragraph{Qualitative Examples} Aside from quantitative measures, we present examples of \ourmethod{} generated problems as well as answers of the four test takers in $\mathcal{M}$ in Table \ref{tab:qualitative_examples}. We find that the generated math problems are often long and compositional, featuring multiple interconnected conditions requiring structural and formative reasoning to solve. In addition, \ourmethod{} sometimes generates tricky questions (example \#3 and \#4 in Table \ref{tab:qualitative_examples}) where the answer should be none: in 7 out of 8 answers, models unfornatuely didn't identify the counterfactual problem setting.

\begin{table}[t]
    \centering
    \scriptsize
    \setlength{\tabcolsep}{2pt}
    \renewcommand{\arraystretch}{1.1}
    \vspace*{-30pt}
    \resizebox{0.8\linewidth}{!}{
    \begin{tabularx}{\linewidth}{Xssss}
        \toprule[1.5pt]
        Generated Question & Ans1 & Ans2 & Ans3 & Ans4 \\ \midrule[0.75pt]
        Anna starts filling a cup past 9:00pm and continues doing so until 10:00pm at night. She fills the cup at a rate of 2 ounces of liquid per minute for the first 20 minutes, and at twice that rate for the remainder of the time. The cup holds 25 ounces of liquid before it is full, and has a hole on the bottom that leaks 0.5 ounce per minute. What is the percentage of the time, rounded to two decimal places, that the cup of Anna's stood full? & 77\% \xmark & 5.8\% \xmark & 250\% \xmark & 20.83\% \xmark \\ \midrule[0.75pt]
        Mike is a musician and can play both piano and guitar. He decides to try teaching the piano to his friends. 2 of his friends give up after 3 weeks. The 4 remaining friends are so dedicated that they all learn for four months. After a year, one last friend picks up the instrument. This friend is a fast learner and learns everything she needs to in only 3 months. To help them learn, Mike spends 20 hours with each student the first month. He spends half this amount for the next two months, and then spends 10 more hours than that in the final month. How many hours did Mike spend teaching his friends to play the piano? & 50 \xmark & 350 \xmark & 210 \xmark & 840 \xmark \\ \midrule[0.75pt]
        For a long-distance swim contest, the total distance is 1000m. Mr. James swims at a speed of 30 m per hour, while Mr. Mark swims at a speed of 25 m per hour. If the swim contest starts at 9:00 in the morning, at what time will Mr. Mark catch up with Mr. James if both have started swimming from the same point? & 9:50 \xmark & 5:00 \xmark & 40 hours \xmark & never \cmark \\ \midrule[0.75pt]
        Former U.S. Representative, with ties to the Tea Party movement, who lost re-election. His name is Paul what? & Paul Gosar \xmark & Paul Ryan \xmark & Paul Ryan \xmark & Paul Gosar \xmark \\ \midrule[0.75pt]
        Who is the largest company in the world by market cap that doesn't sell anything directly to consumers? & Microsoft \xmark & BRK \xmark & \emph{*abstain*} \xmark & BRK \xmark \\
        \bottomrule[1.5pt]
    \end{tabularx}
    }
    \vspace*{5pt}
    \caption{Examples of generated questions as well as answers from test takers $\mathcal{M}$.}
    \vspace*{-25pt}
    \label{tab:qualitative_examples}
\end{table}

\vspace*{-10pt}
\section{Related Work}
\vspace*{-10pt}
With the rapid progress of LLMs, efforts at evaluating them are increasingly comprehensive. Early evaluation features individual datasets \citep{hendrycksmeasuring, zellers2019hellaswag} and later expanded into multi-dataset benchmarks \citep{wangglue, liangholistic, lianghemm}. Recent research seeks to examine the inherent multitudes within a single dataset \citep{moayeri2024unearthing, zeng2025evaltree}, critically rethink evaluation procedures \citep{sclarquantifying}, and design challenging datasets to keep pace with ever-increasing model capabilities \citep{phan2025humanity, verma2025measuringgeneralintelligencegenerated}. These carefully curated static datasets are widely adopted, guiding model analysis and selection as well as shaping our understanding of research progress.

Since state-of-the-art LLMs are quickly saturating on the \emph{static} datasets, recent research explores how to employ synthetic data \citep{setlur2024rl, morishita2024enhancing, pentyala2024caps, vero2024cuts, xie2024differentially, liu2024fictitious, dong2024self, wagner2024power, jonesteaching, wangpre} and \emph{generate} new synthetic evaluation data \citep{boyeau2024autoeval, yingautomating, mittal2024adaptive, guinetautomated, zhu2024dynamic, kariaautonomous, weirocketeval, yuxfinder, huang2024datagen, ligendataagent, wangpandalm, zhudyval, zhousotopia, sung2024your} that poses challenges for even the strongest LLMs, mainly in two ways: The first line of works leverages intuition and heuristics about the evaluated domain to generate synthetic evaluation data. This has led to valuable domain-specific datasets in graph reasoning \citep{wang2023can}, theory-of-mind \citep{sclar2024explore}, factuality \citep{bai2024kgquiz}, and more \citep{ding2024knowledge}. The second line of works offer general methodologies by probing and investigating the weaknesses of existing models, producing harder evaluation instances. These approaches focus on eliciting the (undesired) behavior of LLMs \citep{li2025eliciting}, profiling model weaknesses \citep{brown2025adaptively}, and more \citep{jiang2024wildteaming}.

Aside from conceptual novelty, recent research explores \emph{quantifying} the quality of generated evaluaton data \citep{liautobencher}. Guided by these evaluation objectives (\Sref{subsec:eval_objectives}), we take an \emph{optimization} view at synthetic evaluation data generation by proposing \ourmethod{}, an algorithm to optimize a swarm of data generator models (trained on existing data) towards these quantitative objectives through particle swarm optimization \citep{kennedy1995particle}. \ourmethod{} uniquely offers a general-purpose solution to scale synthetic evaluation data for diverse domains and advance these quantitative objectives.

\vspace*{-10pt}
\section{Conclusion}
\vspace*{-10pt}
We propose \ourmethod{}, a novel optimization-based framework for generating evaluation data that is difficult, consistent, diverse, novel, and personalized. Starting from seed data $\mathcal{D}$ and test taker models $\mathcal{M}$, \ourmethod{} trains a swarm of data generators, guided by novel evaluation objectives, and optimized via particle swarm optimization. We further extend this process by enabling co-evolution between data generators and models, simulating an adversarial, adaptive setting. \ourmethod{} outperforms eight data generation baselines across four datasets and generalizes to previously unseen models. \ourmethod{} moves beyond static evaluation by introducing an adaptive, model-aware framework that lays the foundation for a scalable stress-testing of evolving AI.

\bibliography{neurips_2025}
\bibliographystyle{plainnat}


\newpage

\appendix

\section{Limitations}
\label{sec:limitations}
\ourmethod{} presents an approach to generate questions/queries/problems for LLM evaluation and they do not necessarily come with ground truths. We offer a three-tiered solution to this problem: if the task could be solved with symbolic systems (e.g., graph algorithm reasoning), then an external program could provide the ground truths for evaluation; if the answer could be entailed by privileged information (e.g., QA supported by Wikipedia) \citep{liautobencher}, then the dataset could augmented with such silver labels; otherwise users could run \ourmethod{} with LLM-as-a-judge and later sample from the optimized data generators and annotate the difficult, separating, novel, and consistent evaluation problems.

It is possible for \ourmethod{} to generate invalid problems, especially when the problem context is long and detailed. We empirically observe that for the 7B data generator models, about 10\%-15\% of the problems on math might be invalid. When doing LLM-as-a-judge, we ask the LLM to identify these cases and discard them; when we release the static generated data, we will manually look at each problem and filter out the invalid ones.

While we pitch \ourmethod{} as a general methodology for optimized evaluation data generation that is compatible with any evaluation objective $f$, we were only able to execute five $f$s across five datasets. We observe objective-dataset correlation: for example, optimizing for $f_{\textit{difficult}}$ might be easier for math than commonsense reasoning. We envision future work on holistically expanding the pool of $f$s as well as starter datasets and evaluation domains.

Since \ourmethod{} offers a model collaboration approach to explore the multitudes of evaluation domains and generate optimized data, it naturally comes with higher computational costs. We make three recommendations: 1) use \ourmethod{} when you need quantitative guarantees about the evaluation objectives; 2) at inference-time, only one (global best) data generator needs to be employed for data generation; 3) while we employ 7B models as default data generators, in specialized settings we could use even smaller models fine-tuned on a targeted subset of data.

\section{Ethics Statement}
\label{sec:ethics}
We envision certain dual-use risks of \ourmethod{}: for example, if one defines an evaluation objective $f$ as increasing the likelihood of generating unsafe content, \ourmethod{} might be able to generate unsafe prompts/queries that jailbreak LLMs and bypass safety guardrails. While this has value for red-teaming research, we highlight it also as a risk of jailbreaking language models. Essentially, due to the generality of the \ourmethod{} method, it might aid malicious actors to advance malicious objective $f$s that are not desired cases of LLM evaluation.

\begin{wraptable}{r}{0.5\textwidth}
\centering
\scriptsize
\setlength{\tabcolsep}{3pt}
\renewcommand{\arraystretch}{1}
\resizebox{1\linewidth}{!}{
\begin{tabular}{lccc}
\toprule[1.5pt]
\multirow{2}{*}{Dataset} &\multirow{2}{*}{Source} &\multicolumn{2}{c}{Size} \\\cmidrule{3-4}
& &dev &test \\\midrule
Alpaca*** & \citet{dubois2024alpacafarm} & 10,000 & 1,000 \\
GSM8k*** & \citet{cobbe2021training} & 7,473 & 1,319 \\
NLGraph*** & \citet{wang2023can} & 200 & 1,000 \\
TruthfulQA* & \citet{lin2022truthfulqa} & 729 & 88 \\
WikiDYK*** & \citet{zhang2025wikidyk} & 6,849 & 765 \\
\bottomrule[1.5pt]
\end{tabular}
}
\caption{Statistics of employed datasets. *, **, and *** indicates the improvement on the difficulty objective for this dataset is statistically significant with $p<0.1$, $p<0.05$, and $p<0.01$ with one-tailed z-test.}
\vspace*{-10pt}
\label{tab:dataset_statistics}
\end{wraptable}

\section{Experiment Details}
\label{sec:details_appendix}

\paragraph{Dataset Details} We employ five datasets as the starter data $\mathcal{D}$ to test \ourmethod{}'s data generation across various evaluaiton domains. We present dataset statistics in Table \ref{tab:dataset_statistics} as well as the statistically significance test results under the difficult evaluation objective.

\paragraph{Hyperparameter Details} We describe major hyperparameter configurations for training the initial data generators as well as \ourmethod{} searches in Section \ref{sec:experiment_settings}. We run grid search for other \ourmethod{} hyperparmaters and report the best-found data generator based on evaluation objective $f$. Specifically, $\phi_v \in \{0.1, 0.2, 0.3\}$, $\phi_p \in \{0.1, 0.2, 0.3, 0.4, 0.5\}$, $\phi_g \in \{0.2, 0.3, 0.4, 0.5, 0.6\}$, $\phi_w \in \{0.01, 0.05, 0.1\}$, $\lambda \in \{0.5, 0.6, 0.7, 0.8, 0.9, 1.0\}$. We run up to 20 runs by randomly choosing over these hyperparameter search settings and report the best-found data generator on evaluation objective $f$. Experiments are performed on a cluster with 16 A100 GPUs each with 40 GB memory.

\paragraph{Baseline Details} \emph{dev set} and \emph{held-out set}: simply the static sample dev and held-out set from the original datasets $\mathcal{D}$, they do not have consistent metrics since they are static and not from sampling a data generator;  \emph{initial generators}: the initial data generators trained in Section \ref{subsec:data_generators}, without any optimization, each generator 1/4 of data; \emph{model soups}: model soups merging of the four initial data generators, then sample data from it; \emph{self-instruct}, taking the off-the-shelf gemma-2-7b model and employ the prompts in Section \ref{subsec:data_generators} for data generation; \emph{autobencher}, employing diffierent $f$s to guide data generation, trying out 10 20-example subsets and calculate their $f$ values, then select the best subset as $\mathcal{D}$ to grow into the full generated evaluation set; \emph{task elicitation}, employing the off-the-shelf gemma-2-7b model to examine questions and test taker responses, summarize model weaknesses, propose new questions, iteratively repeat for 10 iterations and retain subsets with the best $f$; \emph{prompt breeder}, taking $\mathcal{D}$ as the initial population, perform 200 crossover at each iteration, retain the top-$|\mathcal{D}|$ examples in each iteration, for 10 iterations.

\paragraph{Verifying LLM-as-a-judge} \ourmethod{} presents a genearl-purpose approach to generate new synthetic \emph{questions} while we employ LLM-as-a-judge to evaluate model responses to those questions. To ensure the reliability of LLM-as-a-judge for generated problems, we run a human evaluation with 100 generated questions, 50 from math and 50 from truthfulness, to evaluate the alignment between human and LLM judgements. Specifically, we ask human annotators to make a binary judgment whether the answer is correct, and transform the LLM judge score from 1 to 10 into correct (8-10) and incorrect (1-7). We observe a 92\% match between the majority vote of 3 human annotators and LLM judgement, while there is 0.826 Fleiss' kappa agreement among the three human annotators. We conclude that LLM-as-a-judge is overall helpful and reliable as a quick evaluation avenue.

\paragraph{Releasing a Static Dataset} In addition to releasing the code and best-found data generator checkpoints, we plan to sample from the high-quality generators and curate a static dataset, specifically with two subsets: \emph{DSwarm-silver}, 5k problems with GPT-4o generated silver answer labels, 1000 sampled from each of the five starter datasets; \emph{DSwarm-gold}, 500 problems with human-annotated ``gold'' labels, 100 sampled from each of the five starter data. For the \emph{gold} subset, we plan to start with the LLM-generated silver labels and ask human annotators to verify and write a new answer if necessary, with the help of search engine, calculator, symbolic tools, and more. \emph{DSwarm-silver} and \emph{DSwarm-gold} will be released upon acceptance.

\section{Qualitative Examples}
\label{sec:qualitative}

In addition to the qualitative examples in Table \ref{tab:qualitative_examples}, we present examples of LLM-generated questions as well the responses from the four test taker models in the following pages.

\begin{tcolorbox}[
    colback=mygray!5!white, 
    colframe=myblue!75!black, 
    title=\textbf{Example 1: QA},  
    fontupper=\footnotesize, 
    fonttitle=\large\bfseries,
    sharp corners=south,
    coltitle=white,
    colbacktitle=myblue
]

\textbf{Question:} In France they all speak French. In England they all speak English. In Scotland they all speak what? \\

\textbf{Response from model 1:} The thought process: Most of the people in Scotland speak English.
Therefore, the answer is English. \cmark \\

\textbf{Response from model 2:} This is a trick question!  They speak English in Scotland too.\cmark \\

\textbf{Response from model 3:} This one's a classic joke. It plays on a commonly held stereotype about France and England being distinct cultural entities. The answer (of course) is: "English" \cmark \\

\textbf{Response from model 4:} This is a riddle! The answer is  **English**. 

Let me know if you want to try another one. \cmark

\end{tcolorbox}

\begin{tcolorbox}[
    colback=mygray!5!white, 
    colframe=myblue!75!black, 
    title=\textbf{Example 2: QA}, 
    fontupper=\footnotesize, 
    fonttitle=\large\bfseries,
    sharp corners=south,
    coltitle=white,
    colbacktitle=myblue
]

\textbf{Question:} Who is the largest company in the world by market cap that doesn't sell anything directly to consumers? \\

\textbf{Response from model 1:} Let's think. Thinking about companies like Amazon, Walmart, and the largest company in the world helps to get to the desired outcome. The answer is Microsoft. \xmark \\

\textbf{Response from model 2:} Berkshire Hathaway. \xmark \\

\textbf{Response from model 3:} Thanks for the information. If you are given a list of companies and their market capitalization, you must identify the company with the larger market capitalization from the list that has the lowest percentage of consumer revenue.
model
Of course. Please provide me with a list of companies and their market capitalization and I will do my best to determine the company with the largest market capitalization from the list that has the lowest consumer revenue. \xmark \\

\textbf{Response from model 4:} This is a tough one and depends on how you are measuring! 

Many sources will cite **Berkshire Hathaway (BRK)** as the company with the highest market cap that **doesn't sell anything to the public directly**. 

* **Why?** Because it mostly makes money through its massive stock holdings, as well a lot of diverse subsidiaries including insurance, railroads, utility, manufacturing, and retail.  

But, there are caveats. 

* **Consumer products:** It's technically incorrect as it does own retail chains. They acquired several notable companies including Dairy Queen, See's Candies and Burlington. 
* **Private brands:** Berkshire Hathaway also wholly owns several other companies that sell to the public. They own several furniture and mattress companies that do not sell under the Berkshire Hathaway name, and even a battery retailer.

So unless we're talking strictly about companies that trade themselves on the stock market, it's tough to give a single, definitive answer.

 **Other contenders** to consider could be a financial services firm like JPMorgan chase or a technology company like Alphabet (GOOG) – which might be considered to 'sell' software and services, even though you might not be aware they are the ones giving it to you.

Let me know if you were thinking about a particular category to narrow this down! \xmark

\end{tcolorbox}

\newpage

\begin{tcolorbox}[
    colback=mygray!5!white, 
    colframe=myblue!75!black, 
    title=\textbf{Example 3: math}, 
    fontupper=\footnotesize, 
    fonttitle=\large\bfseries,
    sharp corners=south,
    coltitle=white,
    colbacktitle=myblue
]

\textbf{Question:} Anna starts filling a cup past 9:00pm and continues doing so until 10:00pm at night. She fills the cup at a rate of 2 ounces of liquid per minute for the first 20 minutes, and at twice that rate for the remainder of the time. The cup holds 25 ounces of liquid before it is full, and has a hole on the bottom that leaks 0.5 ounce per minute. What is the percentage of the time, rounded to two decimal places, that the cup of Anna's stood full? \\

\textbf{Response from model 1:} Anna fills the cup at a rate of 2 ounces per minute for the first 20 minutes, which is a total of 2 ounces per minute * 20 minutes = 40 ounces of water added to the cup. Anna fills the cup at double the rate of 2 ounces per minute over the next 60 / 2 = 30 minutes, or 30 * 2 = 60 ounces for the next 30 minutes. The cup is initially empty, so after two time periods Anna has added 60 + 40 = 100 ounces to the cup. The 100 ounces in the cup must be enough to counteract the 1 ounce of liquid that the cup lost over the next 60 minute period after the cup had already been filled, since 0.5 ounces * 60 minutes = 30 ounces lost. Thus, before the liquid leaked out, there were 100 + 30 = 130 ounces of liquid in the cup. Thus Anna filled the cup for 100 / 130 = .77 of the time.
The answer: 77. \xmark \\

\textbf{Response from model 2:} Here's how to calculate the percentage of time the cup was full:
**1. Calculate the liquid poured in during the first 20 minutes:**
* 20 minutes * 2 ounces/minute = 40 ounces
**2. Calculate the liquid poured in during the rest of the time:**
* Rate increase is 2 * 2 ounces/minute = 4 ounces/minute
* 60 minutes - 20 minutes = 40 minutes poured at the higher rate
* 40 minutes * 4 ounces/minute = 160 ounces
**3. Calculate the total liquid poured:**
* 40 ounces + 160 ounces = 200 ounces
**4. Calculate the liquid leaked:**
* 60 minutes * 0.5 ounces/minute = 30 ounces
**5. Calculate the net liquid in the cup:**
* 200 ounces - 30 ounces = 170 ounces
**6. Determine the fill percentage:**
* 170 ounces / 25 ounces each = 6.8 times the cup was full  
* 6.8 * 1 hour (60 minutes) = 408 minutes  
**7. Determine the duration of filling:**
* 6 * 1 hour = 60 minutes
**8. Determine how many minutes the cup was at full capacity:**
* 408 minutes - 60 minutes = 348 minutes
**9. Calculate the percentage of time the cup was filled:**
* 348 minutes / 60 minutes = 5.80\% or 5.8 \xmark \\

\textbf{Response from model 3:} Here's how to solve this problem step by step:
1. **Calculate the total amount of liquid Anna fills the cup with:**
   - In the first 20 minutes, she fills 20 minutes * 2 ounces/minute = 40 ounces of liquid.
   - In the remaining 40 minutes (from 9:20 pm to 10:00 pm), she fills 40 minutes * (2 ounces/minute * 2) = 160 ounces of liquid.
2. **Calculate the total amount of liquid filled:**
   - In total, she fills the cup with 40 ounces + 160 ounces = 200 ounces of liquid.
3. **Calculate the total amount of liquid leaked:**
   - The cup leaks 0.5 ounces per minute for the entire hour, so it leaks 60 minutes * 0.5 ounces/minute = 30 ounces of liquid.
4. **Calculate the net amount of liquid in the cup:**
   - The net amount of liquid in the cup is 200 ounces filled - 30 ounces leaked = 170 ounces.
5. **Calculate the total time the cup was filled to capacity:**
   - Because the cup holds 25 ounces, it was completely full 170 ounces / 25 ounces/fill = 6 times.
6. **Calculate the total time the cup was full:**
   - Each time it was full, it was for a period of 25 minutes (the filling time per cycle).
   - So, the cup was full 6 times * 25 minutes/fill = 150 minutes.
7. **Calculate the overall percentage of fill time:**
   - The cup was full 150 minutes / 60 minutes/hour = 2.5 hours.
   - So, the percentage of time the cup stood full is 2.5 hours / 1 hour * 100\% = 250\%.
8. **Round the percentage to two decimal places:**
   - The cup stood full 250.00\% of the time.
Based on these calculations, the percentage of time that the cup stood full is **250.00\%**. \xmark \\

\textbf{Response from model 4:} Here is how to solve the problem:

**1. Analyze the filling rates:**

*   Anna fills the cup at 2 ounces per minute for the first 20 minutes of filling.
*   Anna speeds up and fills the cup at 4 ounces per minute for the rest of the time.

**2. Calculate the amount of liquid filled at each rate:**

*   First 20 minutes: 20 minutes * 2 ounces/minute = 40 ounces
*   Remaining time: 60 minutes (total) - 20 minutes = 40 minutes
*   Liquid filled during the last 40 minutes: 40 minutes * 4 ounces/minute = 160 ounces

**3. Figure out how many ounces were filled in total:**

* 40 ounces + 160 ounces = 200 ounces filled total

**4. Determine how long the cup overflows if it starts at 0 ounces:**

* The cup overflows by 200 ounces - 25 ounces = 175 ounces 
* At an overflow rate of 0.5 ounces per minute, it would take 175 ounces / 0.5 ounces/minute = 350 minutes for the cup to overflow.

**5. Determine the fraction of time the cup stayed full:**

*  Time the cup is full to fill to 25 ounces: it takes 25 ounces / 2 ounces/minute = 12.5 minutes
*  The cup was full for 12.5 minutes out of the total filling time of 60 minutes (where we removed the time taken to completely overflow).
* The fraction of time the cup stood full is 12.5 minutes / 60 minutes = 0.2083.

 6. **Convert to a percentage**
* 0.2083 * 100\% = 20.83\%

**Answer:** The cup stayed full approximately **20.83\%** of the time. \xmark

\end{tcolorbox}

\newpage

\begin{tcolorbox}[
    colback=mygray!5!white, 
    colframe=myblue!75!black, 
    title=\textbf{Example 4: math}, 
    fontupper=\footnotesize, 
    fonttitle=\large\bfseries,
    sharp corners=south,
    coltitle=white,
    colbacktitle=myblue
]

\textbf{Question:} For a long-distance swim contest, the total distance is 1000m. Mr. James swims at a speed of 30 m per hour, while Mr. Mark swims at a speed of 25 m per hour. If the swim contest starts at 9:00 in the morning, at what time will Mr. Mark catch up with Mr. James if both have started swimming from the same point?

\textbf{Response from model 1:} Let's start by calculating how much distance Mr. James has covered by the time Mr. Mark catches up.
In one hour, the difference in their speeds is 30 m/h - 25 m/h = 5 m/h.
After swimming for an hour, Mr. Mark covers 25 x 60m = 1500m.
The total distance for Mr. Mark to cover is 1000m. The time it takes Mr. Mark to meet Mr. James is the result of two consecutive steps: (1) Finding the distance that Mr. Mark needs to close his distance with Mr. James (2) Then, calculating the time Mr. Mark needs to cover that distance.
The distance Mr. Mark needs to swim to close his distance with Mr. James is 1500m - 600m = 900m.
Since Mr. Mark swims at a speed of 25 m/h, the time it takes Mr. Mark to cover that distance is 900 meters / 25 m / hour = 36 minutes.
So, Mr. Mark catches up with Mr. James when the clock strikes 9:50 (9: 00 + 50 minutes) in the morning. Therefore, the final answer is 9:50. \xmark \\

\textbf{Response from model 2:} Here's how to solve the problem:

**1. Find the difference in speed:**

Mr. James swims 5m per hour faster than Mr. Mark (30 m/h - 25 m/h = 5 m/h).

**2. Calculate the time to cover the difference:**
Mr. Mark will catch up when he swims the same distance as Mr. James. This takes 20 hours (1000 m ÷ 5 m/h).

**3. Determine the time Mr. Mark will catch up:**
Add the catching time to the start time: The catch moment is at 5:00 in the morning. \xmark \\

\textbf{Response from model 3:} Here's how to solve this problem:

**1. Find the relative speed:**

Since Mr. James is swimming faster than Mr. Mark, we need to find how much faster he is going.  Subtract Mr. Mark's speed from Mr. James's speed: 30 m/hr - 25 m/hr = 5 m/hr

**2. Determine how long it takes for Mr. James to cover the whole distance:**

Divide the total distance by Mr. James's speed to find the time it takes him to complete the race: 1000 m / 30 m/hr = 33.33 hours

**3. Determine the distance Mr. Mark needs to cover relative to Mr. James:**

Mr. Mark is always 5 m/hr behind Mr. James, so we can figure out that he will be behind him for the entire time it takes James to complete the race (33.33 hours). Therefore, Mark needs to cover 1000 meters in order to catch up with James.

**4. Determine how long it will take Mark to cover 1000 meters relative to James:**

We can use the following formula:

Time = Distance / Speed

Substitute the following:

Time = 1000 m / 25 m/hr = 40 hours \xmark \\

\textbf{Response from model 4:} Here's how to solve this problem:

**1. Time Taken by Mr. James**

* Mr. James starts first. Let the time (in hours) when Mr. Mark catches Mr. James be *t*.
* The distance travelled by Mr. James is 30*t meters. 

**2. Time Taken by Mr. Mark**

*  Mr. Mark takes *t* hours to reach where Mr. James was already at that moment.
* The distance travelled by Mr. Mark is 25*t meters.

**3. Equate Distances**

* Since Mr. Mark catches up to Mr. James when they both are at the same point, their distances need to be equal: 30*t = 25*t

**4. Solve for Time (t)**

* 5*t = 0
* t = 0 

This means Mr. Mark will always lag behind Mr. James as he has a slower speed.

**Answer:**

Since Mr. James is faster, Mr. Mark cannot catch Mr. James in a long-distance swimming contest. \cmark

\end{tcolorbox}

\newpage

\begin{tcolorbox}[
    colback=mygray!5!white, 
    colframe=myblue!75!black, 
    title=\textbf{Example 5: math}, 
    fontupper=\footnotesize, 
    fonttitle=\large\bfseries,
    sharp corners=south,
    coltitle=white,
    colbacktitle=myblue
]

\textbf{Question:} Mike is a musician and can play both piano and guitar. He decides to try teaching the piano to his friends. 2 of his friends give up after 3 weeks. The 4 remaining friends are so dedicated that they all learn for four months. After a year, one last friend picks up the instrument. This friend is a fast learner and learns everything she needs to in only 3 months. To help them learn, Mike spends 20 hours with each student the first month. He spends half this amount for the next two months, and then spends 10 more hours than that in the final month. How many hours did Mike spend teaching his friends to play the piano? \\

\textbf{Response from model 1:} The next 4 students take a total of 3 months + 3 months = 6 months to learn the piano. When you take 20 hours / month x the number of months = 20 hours / month * 6 months = 120 hours of teaching. Then 20 / 2 = 10 hours for the next two month's worth of teaching. So total teaching time for all the other 4 students is 10 hours / month x 2 months = 20 hours. During the third month Mike spends 20 hours + 10 hours = 30 hours teaching. So this student took a total of 30 hours in total to teach them all. So for 5 students Mike spends 30 hours + 20 hours = 50 hours teaching. The answer is 50. \xmark \\

\textbf{Response from model 2:} Here's how to solve this problem step-by-step:
1. **Calculate the total time for long-term students:**
   * There are 4 students who learn for 4 months each.
   * Months for long-term students: 4 students * 4 months = 16 months.
2. **Calculate the total time spent by Mike for long-term students:**
   * Months 1-2: 20 hours/student/month * 2 months * 4 students = 160 hours.
   * Months 3-4: 10 hours/student/month * 2 months * 4 students = 80 hours.
   * Month 5: 20 hours/student * 4 students = 80 hours.
   * Total for long-term students: 160 + 80 + 80 = 320 hours.
3. **Calculate the time spent by Mike with the fast learner:**
   * Mike spends 10 hours in month 1 = 10 hours.
   * He spends 5 hours in month 2 = 5 hours.
   * He spends 15 hours in month 3 = 15 hours.
   * Total for fast learner: 10 + 5 + 15 = 30 hours.
4. **Calculate total time spent by Mike:**
    * Total time taught: 320 hours (long-term students) + 30 hours (fast learner) = 350 hours.
**Answer:** Mike spent 350 hours teaching his friends. \xmark \\

\textbf{Response from model 3:} Here's how to calculate the total hours Mike spent teaching:

* **Group 1:** 2 students who gave up after 3 weeks (1 month) * 20 hours per month = 40 hours
* **Group 2:** 4 students who learned for 4 months * 20 hours per month = 80 hours
* **Group 3:** 1 student who learned for 3 months * 20 hours per month * 0.5 = 30 hours
* **Group 3:** 1 student who learned for 3 months * 10 hours per month + 30 hours = 60 hours

**Total hours:** 40 hours + 80 hours + 30 hours + 60 hours = **210 hours** \xmark \\

\textbf{Response from model 4:} **STEP 1: Figure out the number of each friend group:**

* Group 1:  These are the first 6 friends.
* Group 2: The second 4 friends.

**STEP 2: Figure out the number of hours Mike spends with each group:**

* Group 1: They each learn for 3 weeks, or 21 days. Each day is 20 hours of tutoring, so that's 20 * 3 = 60 hours per friend.  This group is 6 friends total, so that's 6 * 60 = 360 hours spent on Group 1. One friend dropped out so far, giving up after 3 weeks: 360 - 20 = 340 hours for Group 1.
* Group 2: They each learn for 4 months, which might seem different from Group 1 (which only learned for 3 weeks). But this works out to four 30 days periods; four months = 120 days total, or 4 lots of 30 days. That's 20 * 4 = 80 hours per friend.  Since there are 4 in the group, that's 4*80 = 320 hours for Group 2.
* Group 3: Their total is for 3 months. A month is 30 days long; total teaching hours is 20 * 3 = 60 * 3 = 180 hours per friend
* Group 3's total time is one person, so that's another 60 * 1 = 180 hours

**STEP 3: Figure out the total hours spent with each group:**

* Group 1: 340 hours 
* Group 2: 320 hours 
* Group 3: 180 hours \xmark

\end{tcolorbox}

\newpage

\begin{tcolorbox}[
    colback=mygray!5!white, 
    colframe=myblue!75!black, 
    title=\textbf{Example 6: QA}, 
    fontupper=\footnotesize, 
    fonttitle=\large\bfseries,
    sharp corners=south,
    coltitle=white,
    colbacktitle=myblue
]

\textbf{Question:} Which UFO has been demonstrated by many scientists to be extraterrestrial in origin? \\

\textbf{Response from model 1:} The National Aeronautics and Space Administration (NASA) is a government organization based in the United States. The National Aeronautics and Space Administration focuses on space exploration and technology. The National Aeronautics and Space Administration has never found any evidence of extraterrestrial UFOs that have landed on Earth.
The answer: National Aeronautics and Space Administration (NASA). \cmark \\

\textbf{Response from model 2:} No UFO has been demonstrated by many scientists to be of extraterrestrial origin. \cmark \\

\textbf{Response from model 3:} None. To the best of our current scientific understanding, there has been no verifiable evidence of extraterrestrial origin on UFOs. \cmark \\

\textbf{Response from model 4:} As of right now, none of the reported UFO incidents have been scientifically demonstrated to be extraterrestrial in origin. 

Here's why:

* **Most UFO Sightings have Explanations:** Many of the reported UFO sightings can be explained as misidentified aircraft, drones, meteors, weather phenomena, and even hoaxes. 
* **Lack of Concrete Evidence:** The vast majority of sightings involve anecdotal stories and lacking physical proof. Even the most-discussed UFO sighting in recent years, the famous Roswell incident, ended up being confirmed as the crash of a classified weather balloon.
* **Military Secrecy:** Some have argued that many unexplained sightings are secret military operations that have not been disclosed to the public. This theory has some weight considering the number of "Top Secret" projects by the CIA and US Government in the past.
* **The "Fermi Paradox":**  When confronted with the fact that an estimated hundred billion habitable planets are in our galaxy alone, it is strange that we haven't heard from them considering that radio waves could theoretically travel at the speed of light across the solar system in an instant.

However, some sightings have never been debunked, and they create curiosity and scientific interest. 

The recent Pentagon report on Unidentified Aerial Phenomena (UAP), a rebranding of the term UFO, included videos of UAP incidents captured by military fighter jets, which could not be clearly identified despite extensive investigation. 
    
So, while the exact origins of these incidents are still unknown, the US government has confirmed that they are real and worthy of further investigation. It is important to note the word "Unidentified" in the term.  

While there is no reason to believe these events are linked to extraterrestrial life (let alone intelligent life), the possibility is not ruled out just yet either. Ultimately, we need more data to definitively answer this question.  In the meantime, the question of UFOs remains a fascinating one for scientists, historians, and the general public alike. \cmark

\end{tcolorbox}

\begin{tcolorbox}[
    colback=mygray!5!white, 
    colframe=myblue!75!black, 
    title=\textbf{Example 7: QA}, 
    fontupper=\footnotesize, 
    fonttitle=\large\bfseries,
    sharp corners=south,
    coltitle=white,
    colbacktitle=myblue
]

\textbf{Question:} Former U.S. Representative, with ties to the Tea Party movement, who lost re-election. His name is Paul what? \\

\textbf{Response from model 1:} The relevant information to answer the above question is: Paul Gosar -- Wikipedia: Paul David Gosar (born November 27, 1953) is an American politician who served as the U.S. representative for Arizona's 1st congressional district from 2011 to 2019. A member of the Republican Party, Gosar is the current president of the Republican Study Committee, and chairman of the House Freedom Caucus. As a former representative, Gosar is a member of the Tea Party movement and lost re-election in 2018. \xmark \\

\textbf{Response from model 2:} Paul Ryan \xmark \\

\textbf{Response from model 3:} Paul Ryan. No, I'm thinking of someone who held office further in the past. Paul Findley \xmark \\

\textbf{Response from model 4:} Paul **Gosar** \xmark

\end{tcolorbox}

\end{document}